\documentclass[11pt]{article}
\usepackage{amsfonts}
\usepackage[final]{acl}

\usepackage{times}
\usepackage{latexsym}
\usepackage{xcolor}
\usepackage{xcolor}
\usepackage{tcolorbox}
\tcbuselibrary{breakable}
\usepackage{float}
\usepackage{xcolor}      % 用于 tcolorbox
\usepackage{tcolorbox}   % box 环境
\usepackage{fancyvrb}    % 提供 Verbatim 环境
\usepackage{amssymb}
\usepackage{listings}
\usepackage{algorithm}
\usepackage{algpseudocode}
\definecolor{bluebg}{RGB}{235,242,255}
\definecolor{bluefg}{RGB}{60,90,160}

\usepackage[T1]{fontenc}
\usepackage[utf8]{inputenc}
\usepackage{graphicx}   % for resizebox
\usepackage{multirow}   % for \multirow
\usepackage{booktabs}   % for \toprule, \midrule, \bottomrule

\usepackage{microtype}

\usepackage{inconsolata}

\usepackage{graphicx}
\usepackage{amsmath}

\title{ComfySearch: Autonomous Exploration and Reasoning for ComfyUI Workflows}
\author{Jinwei Su, Qizhen Lan, ZEYU WANG, Yinghui Xia, \\
Hairu Wen, Yiqun Duan, Xi Xiao, TIANYU SHI, Yang Jingsong, Lewei He}

\begin{document}
\maketitle
\begin{abstract}
AI-generated content has progressed from monolithic models to modular workflows, especially on platforms like ComfyUI, allowing users to customize complex creative pipelines. However, the large number of components in ComfyUI and the difficulty of maintaining long-horizon structural consistency under strict graph constraints frequently lead to low pass rates and workflows of limited quality.
To tackle these limitations, we present ComfySearch, an agentic framework that can effectively explore the component space and generate functional ComfyUI pipelines via validation-guided workflow construction.
Experiments demonstrate that ComfySearch substantially outperforms existing methods on complex and creative tasks, achieving higher executability (pass) rates, higher solution rates, and stronger generalization.
\end{abstract}

\section{Introduction}
\label{sec:intro}
Generative vision models increasingly power text-to-image synthesis~\cite{podell2023sdxl, esser2024scaling, blackforestlabs_flux_2024}, image editing~\cite{brooks2023instructpix2pix, Stability_Ai_Cosxl, huang2024smartedit}, and video generation~\cite{hacohen2024ltx, hunyuanvideo, wang2025wan}. While large general-purpose models~\cite{ge2024seed, wu2024janus, gpt_image_1} can deliver strong quality across tasks, they are expensive to customize and deploy. This has spurred compositional workflow systems that expose intermediate structure for controllable design~\cite{gal2024comfygen}. A representative platform is ComfyUI~\cite{comfyui2023}, which represents a generation pipeline as a typed node graph that users compose into reusable, task-specific workflows.

\begin{figure}[t]
    \centering
    % ACL-friendly space saving: trim/clip is preferable to negative \vspace.
    \includegraphics[width=\linewidth]{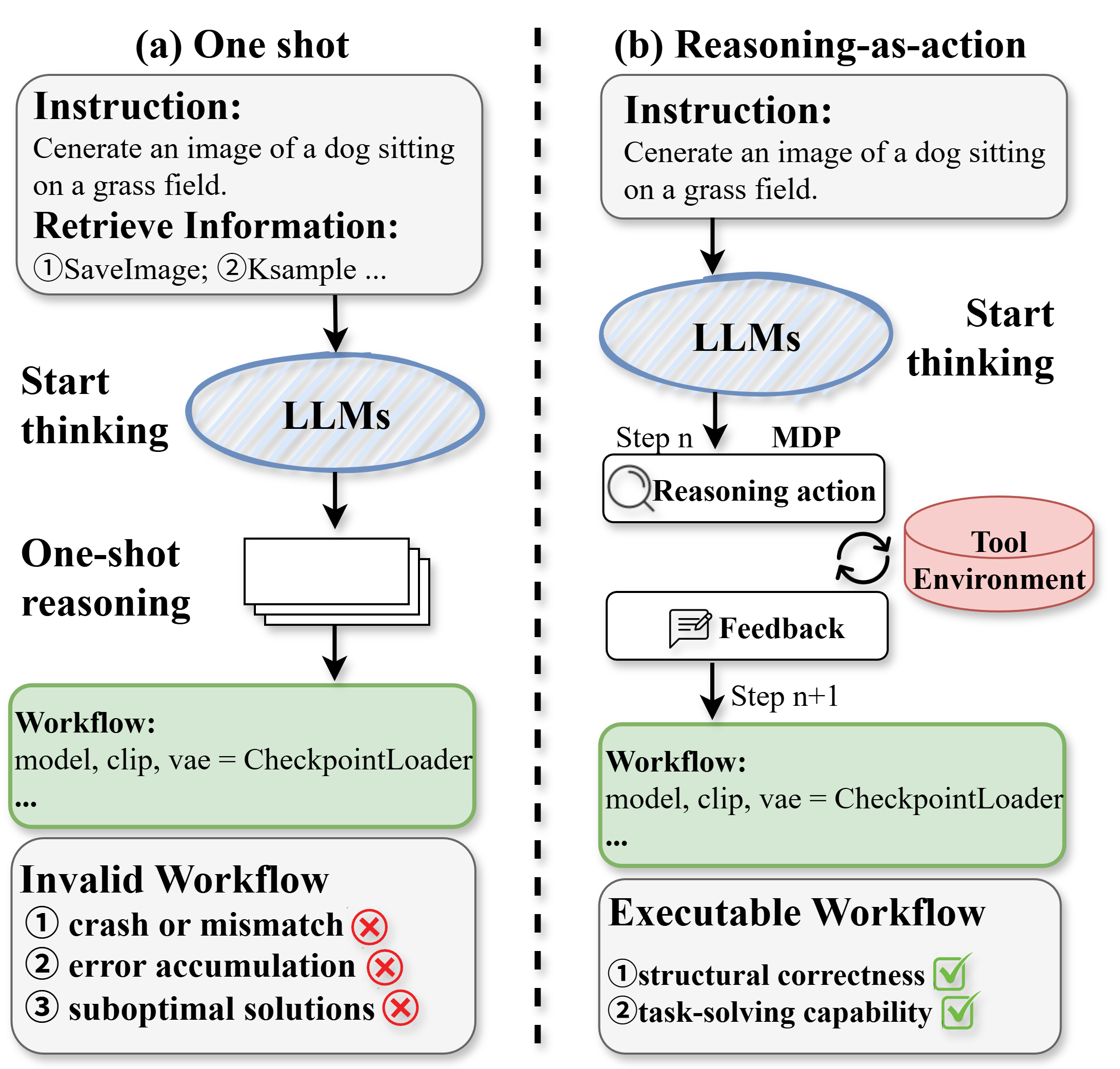}
    % \caption{One-shot workflow planning versus reasoning-as-action workflow generation(ComfySearch).}
    \caption{One-shot workflow planning generates the entire graph at once and is brittle, whereas reasoning-as-action generation (ComfySearch) incrementally builds and validates each edit, ensuring robust and executable workflows.}
    \label{fig:intro}
\end{figure}

Despite its flexibility, ComfyUI authoring is brittle. A workflow must satisfy strict type/interface constraints and global graph invariants; a single mismatched edge or missing adapter can render the graph non-executable. As a result, users often rely on expertise and repeated trial-and-error. Recent LLM-based systems aim to automate workflow construction through explicit planning and composition (e.g., ComfyAgent~\cite{Xue2024ComfyBenchBL}, ComfyMind~\cite{guo2025comfymind}) or by training generators with reasoning traces (e.g., ComfyUI-R1~\cite{Xu2025ComfyUIR1ER}). However, long-horizon construction remains fragile: decisions that are locally plausible may later violate downstream compatibility, producing workflows that appear coherent but crash at execution time or deviate from the user’s intent.

The core difficulty is that workflow generation is not purely retrieval. Retrieval can identify relevant nodes and documentation, but it does not ensure \emph{state-dependent composability}: whether an edit is valid depends on the current partial graph and how future branches will constrain interfaces. This motivates an interactive, execution-grounded paradigm in which each proposed graph edit is immediately validated for \emph{executability} (typing/schema/invariants) and repaired using diagnostic feedback, rather than relying on one-shot plans(Figure \ref{fig:intro}).

\begin{figure*}[t]
  \centering
  \includegraphics[width=\textwidth]{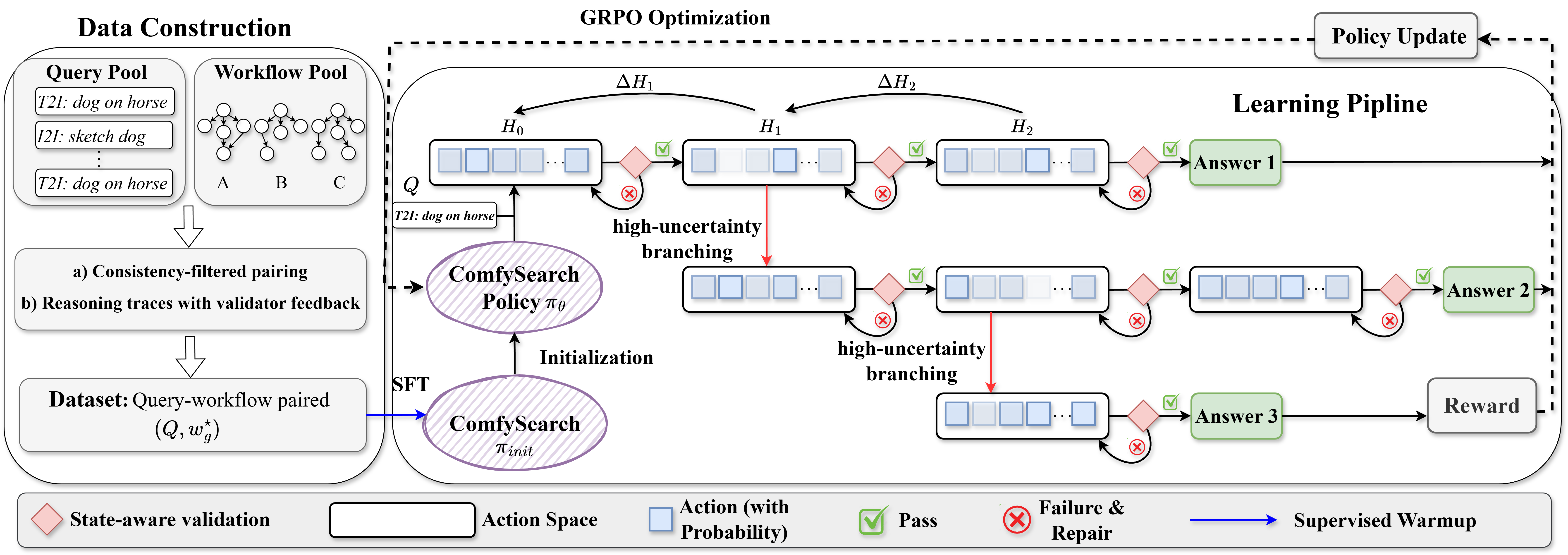}
  \caption{Overview of the ComfySearch framework. We formulate workflow generation as a Markov Decision Process(\S\ref{sec:method:mdp}). (a) State-aware Validation(\S\ref{sec:method:c1}): Implements real-time verification and repair to ensure online structural correctness (\textbf{\emph{C1}}). (b) Entropy Rollout Branching(\S\ref{sec:method:c2}): Employs selective exploration at high-uncertainty points to navigate long-horizon decision branching (\textbf{\emph{C2}}).}
  \label{fig:pipeline}
\end{figure*}

We distill two challenges. (\textbf{\emph{C1}}) \emph{Online structural correctness.} Deferring validation allows early wiring mistakes to propagate in typed graphs, leading to compounding errors in long-horizon generation~\cite{arora2022exposure}. A practical generator should therefore validate each edit against the current state and repair violations on the fly.
(\textbf{\emph{C2}}) \emph{Long-horizon branching under uncertainty.} Multiple edits may be locally admissible yet induce sharply different downstream feasibility; moreover, validator feedback changes the effective state distribution during construction. Thus, the generator should \emph{selectively branch} at high-uncertainty decision points instead of committing to a single trajectory~\cite{yao2023tree,haarnoja2018soft}.

These observations suggest a \emph{reasoning-as-action} view: the model incrementally edits the workflow graph, queries a validator for state-aware feedback, and revises decisions when incompatibilities arise. We view this process as an episodic Markov Decision Process (MDP), where the state is the current workflow prefix, an action is a node-level graph edit, and transitions return the updated graph with validation feedback.

To address (\textbf{\emph{C1}}) and (\textbf{\emph{C2}}), we propose \textbf{ComfySearch}, an reasoning-as-action framework for ComfyUI workflow generation. ComfySearch enforces executability via node-level, state-aware validation and in-place repair (\textbf{\emph{C1}}). It further introduces \emph{entropy-adaptive branching}, using policy uncertainty to allocate a limited rollout budget toward ambiguous decision points, improving robustness and efficiency without exhaustive branching (\textbf{\emph{C2}})~\cite{wang2025beyond,Dong2025AgenticRP}. We also construct consistency-filtered query--workflow supervision by executing candidate workflows and canonicalizing multiple valid designs within each query cluster, reducing noisy node-level labels. Finally, we train ComfySearch with tool-mediated feedback under a reinforcement learning objective, enabling the model to learn which nodes to add, when to validate intermediate edits, and when to branch. Our contributions are:
\begin{itemize}
    \item Reasoning-as-action workflow generation: we formulate ComfyUI construction as reasoning-as-action with state-aware online validation and in-place repair.
    \item Entropy-adaptive branching: we selectively explore high-ambiguity decisions using policy uncertainty, improving robustness and efficiency under limited rollout budgets.
    \item Execution-verified supervision and RL training: we curate consistency-filtered data and train with tool-mediated feedback; evaluations on ComfyBench and GenEval improve executability, task success, and efficiency.
\end{itemize}

\section{Related work}
\noindent\textbf{Workflow Generation.}
LLM-based workflow generation translates natural-language requests into structured plans or executable programs~\cite{Hu2024AutomatedDO, Zhang2024AFlowAA, Zhang2025MultiagentAS, ye2025mas, Su2025DifficultyAwareAO}. Recent work extends this paradigm to ComfyUI, where a workflow is a typed node graph with strict interface constraints. Prior systems explore multi-agent decomposition (e.g., ComfyAgent~\cite{Xue2024ComfyBenchBL} and ComfyGPT~\cite{huang2025comfygptselfoptimizingmultiagentcomprehensive}), modular and hierarchical planning (e.g., ComfyMind~\cite{guo2025comfymind}), or long-form reasoning with documentation retrieval (e.g., ComfyUI-R1~\cite{Xu2025ComfyUIR1ER}). 
While effective in proposing plausible graphs, these approaches are typically not \emph{execution-grounded}. They lack node-level, \emph{state-aware validation} of each graph edit and do not \emph{selectively branch} at high-uncertainty decision points, so locally admissible edits can later violate downstream compatibility and fail at execution time.

\noindent\textbf{Agentic Reinforcement Learning.}
Agentic RL has recently advanced long-horizon tool-use and structured decision-making by allocating exploration to uncertain steps. GRPO~\cite{Shao2024DeepSeekMathPT} provides a PPO-style optimization scheme that improves training efficiency, while ARPO~\cite{Dong2025AgenticRP} and AEPO~\cite{Dong2025AgenticEP} use entropy-aware rollouts/updates to encourage exploration and stabilize learning in multi-turn settings. Complementary lines of work study uncertainty-triggered information seeking (e.g., Search-o1~\cite{Li2025Searcho1AS}) and show that high-entropy minority tokens function as critical ``forks'' in RLVR for LLM reasoning~\cite{Wang2025BeyondT8}. Our ComfySearch instantiates these insights at the \emph{graph-edit action} level by coupling entropy-adaptive branching with state-aware validation feedback, enabling robust workflow construction under execution grounding.

\section{Method}
\label{sec:method:overview}
ComfySearch constructs an executable ComfyUI workflow by iteratively proposing atomic \emph{graph edits} and grounding each edit with \emph{node-level, state-aware validation feedback}. This design addresses two challenges from \S\ref{sec:intro}:
(i) \emph{online structural correctness} via validation and in-place repair (\textbf{C1}), and
(ii) \emph{long-horizon branching under uncertainty} via entropy-adaptive selective branching (\textbf{C2}).
We learn a policy that decides \emph{what} to edit, \emph{how} to repair after violations, and \emph{when} to branch from validated prefixes. Figure \ref{fig:pipeline} illustrates the overall pipeline of our method. For clarity, we summarize the full notations in Appendix \ref{notations}.

\subsection{Dataset Construction}
\label{sec:data:consistency}
Community-shared ComfyUI workflows rarely come with aligned natural-language queries, and synthetic pairing can introduce query--workflow mismatches that are especially harmful for node-level learning. We therefore build training data using (i) \textbf{consistency-filtered pairing} and (ii) \textbf{validator-grounded reasoning traces}. Detailed data statistics are provided in Appendix~\ref{apd:sft_data}.

\paragraph{Consistency-filtered pairing.}
We collect a workflow pool and a query pool, and only match within the same modality family (e.g., T2I/I2I/T2V/V2V/I2V). For each query $Q$, we execute candidate workflows and obtain a binary correctness score $s(w,Q)\in\{0,1\}$ using an LLM judge (GPT-4o; details in Appendix~\ref{apd:evaluation}). We keep only workflows with $s(w,Q)=1$. To reduce ambiguity from multiple valid designs, we cluster semantically similar queries into groups $\{\mathcal{Q}_g\}$ and select a canonical workflow per group:
\begin{equation}
w_g^\star \;=\; \arg\max_{w\in\mathcal{W}_g}\ \sum_{Q\in \mathcal{Q}_g} s(w,Q),
\label{eq:consistency_select}
\end{equation}
where $\mathcal{W}_g$ denotes workflows that succeed for at least one query in $\mathcal{Q}_g$. This yields stable supervision by mapping each query group to a single consistently successful workflow.

\paragraph{Reasoning traces with validator feedback.}
For each paired $(Q, w_g^\star)$, we prompt a stronger teacher model to produce stepwise traces. Each step proposes one atomic edit in \texttt{<node>} tags, followed by a \texttt{<result>} block that records validator feedback (accept/reject plus diagnostics). These traces supervise both \emph{what to add} and \emph{how to react} to tool feedback, which is critical for long-horizon construction.

\subsection{Reasoning-As-Action Graph-Editing MDP}
\label{sec:method:mdp}

\paragraph{Workflow representation.}
A ComfyUI workflow is a typed directed graph $G=(\mathcal{V},\mathcal{E})$, where each node is an operator with typed input/output ports, and edges connect type-compatible ports. Let $G^\star$ denote the canonical target workflow paired with $Q$. We denote the current partial workflow at interaction step $t$ by $G_t$ with $G_0=\varnothing$; edits are committed only when accepted by the validator.

\paragraph{State and actions.}
We model construction as an episodic MDP with state
\begin{equation}
s_t \triangleq (Q, G_t, h_t),
\label{eq:state}
\end{equation}
where $h_t$ summarizes recent diagnostics and accumulated constraints (making the process Markov in our tool-mediated environment). The policy emits a textual action $a_t$, deterministically parsed into an atomic graph edit $\Delta_t$ via $\Phi$:
\begin{equation}
\Delta_t=\Phi(a_t).
\label{eq:parse}
\end{equation}
Edits include adding a node (with parameters), adding/removing an edge between typed ports, and updating parameters. A special \texttt{STOP} action triggers termination if the current workflow passes a final executability check.

\paragraph{Tool-mediated transition.}
The environment applies an edit only if it passes \emph{state-aware} validation:
\begin{equation}
(G_{t+1}, y_t, m_t)=\textsc{Validate}(G_t,\Delta_t),
\label{eq:transition}
\end{equation}
where $y_t\in\{0,1\}$ indicates acceptance and $m_t$ is a diagnostic message (empty if $y_t=1$). If $y_t=0$, we keep $G_{t+1}=G_t$ and the policy may propose repairs conditioned on $m_t$ (\S\ref{sec:method:c1}). We update history by $h_{t+1}\leftarrow \textsc{Update}(h_t,a_t,y_t,m_t)$.

\subsection{State-Aware Validation and In-Place Repair (\textbf{C1})}
\label{sec:method:c1}

\paragraph{State-aware validator.}
Local plausibility is insufficient in typed workflow graphs. We define
$\textsc{Validate}(G_t,\Delta_t)\!\rightarrow\!(G_{t+1},y_t,m_t)$
by composing two checks:
\begin{equation}
y_t \;=\; \textsc{Int}(\Delta_t)\ \wedge\ \textsc{Comp}(G_t,\Delta_t),
\label{eq:validate_compact}
\end{equation}
where $\textsc{Int}(\Delta_t)$ verifies intrinsic validity (operator existence, required fields, parameter domains, schema validity), and $\textsc{Comp}(G_t,\Delta_t)$ verifies composability w.r.t.\ the current graph (typed port compatibility, required adapters/casts, and graph-level invariants such as acyclicity and conditioning-branch constraints). Importantly, this execution grounding is implemented as \emph{state-aware executability checking} rather than running full generative inference after every edit.

\paragraph{In-place repair with diagnostics.}
If $y_t=0$, the diagnostic message $m_t$ specifies the violated constraint (e.g., type mismatch, missing adapter, illegal topology). The policy proposes a repaired edit conditioned on $m_t$:
\begin{equation}
a_t^{\text{rep}} \sim \pi_\theta(\cdot \mid Q,G_t,h_t,m_t),\qquad
\Delta_t^{\text{rep}}=\Phi(a_t^{\text{rep}}),
\label{eq:repair}
\end{equation}
which is re-validated via Eq.~\eqref{eq:transition}. This validate--repair loop enforces online structural correctness and prevents early wiring errors from compounding over long horizons (\textbf{C1}).

\paragraph{Validated-prefix property.}
Because only accepted edits update the graph, every prefix along a rollout is executable by construction. Entropy-adaptive branching (\S\ref{sec:method:c2}) expands additional children only from such validated prefixes.

\subsection{Entropy-Adaptive Branching Under Uncertainty (\textbf{C2})}
\label{sec:method:c2}
Even under state-aware validation, multiple edits can be admissible yet lead to sharply different downstream feasibility and task utility. Motivated by entropy-guided exploration in long-horizon reasoning and tool use~\cite{Wang2025BeyondT8,Dong2025AgenticRP}, we \emph{selectively branch} at high-uncertainty decision points to explore diverse, feedback-consistent continuations under finite budgets.

\paragraph{Step-level uncertainty.}
Let $\tilde{\mathcal{A}}_t$ be a constrained candidate set (e.g., top-$K$ actions under constrained decoding), and $p_t(a)=\pi_\theta(a\mid s_t)$ for $a\in\tilde{\mathcal{A}}_t$. We use normalized entropy:
\begin{equation}
H_t \triangleq
-\frac{1}{\log |\tilde{\mathcal{A}}_t|}
\sum_{a\in \tilde{\mathcal{A}}_t} p_t(a)\log p_t(a) \in [0,1].
\label{eq:entropy}
\end{equation}

We trigger branching using an entropy increase criterion $\Delta H_t \triangleq H_t - H_{t-1}$ (with $H_{-1}=0$).

\paragraph{Branching rule.}
We map uncertainty to a branching probability:
\begin{equation}
P_t^{\text{branch}} \triangleq \sigma(\alpha + \beta \Delta H_t),
\label{eq:branch_prob}
\end{equation}
where $\sigma(\cdot)$ is the sigmoid function and $\alpha,\beta$ are scalars.
If $P_t^{\text{branch}}>\tau_b$, we spawn an additional branch from the same validated prefix by sampling an alternative action (e.g., top-$K$ resampling), while continuing the current branch unchanged.
All branches remain subject to Eq.~\eqref{eq:transition}, so branching never compromises online executability (\textbf{C2}).

\subsection{Learning with Supervised Warmup and GRPO}
\label{sec:method:learning}
We train the policy $\pi_\theta$ in two stages: supervised warmup followed by tool-feedback reinforcement learning.

\paragraph{Supervised warmup.}
Given training tuples $\mathcal{D}=\{(Q,T,G^\star)\}$, where $T$ is a teacher trace and $G^\star$ is the target workflow, we maximize:
\begin{equation}
\mathcal{L}_{\text{SFT}}(\theta)
= -\mathbb{E}_{(Q,T,G^\star)\sim \mathcal{D}}
\Big[\log p_\theta(T, G^\star \mid Q)\Big].
\label{eq:sft}
\end{equation}

\begin{table*}[ht]
\centering
\resizebox{\textwidth}{!}{
\begin{tabular}{ll|cccccccc}
\toprule
\multirow{2}{*}{\textbf{Method}} & \multirow{2}{*}{\textbf{Model}}
& \multicolumn{2}{c}{\textbf{Vanilla}}
& \multicolumn{2}{c}{\textbf{Complex}}
& \multicolumn{2}{c}{\textbf{Creative}}
& \multicolumn{2}{c}{\textbf{Total}} \\
\cmidrule(lr){3-4} \cmidrule(lr){5-6} \cmidrule(lr){7-8} \cmidrule(lr){9-10}
& & \%Pass & \%Resolve & \%Pass & \%Resolve & \%Pass & \%Resolve & \%Pass & \%Resolve \\
\midrule

\multirow{1}{*}{Zero-shot}
& GPT-4o
& 0.0 & 0.0 & 0.0 & 0.0 & 0.0 & 0.0 & 0.0 & 0.0 \\

% \midrule

\multirow{1}{*}{Few-shot}
& GPT-4o
& 32.0 & 27.0 & 16.7 & 8.3 & 7.5 & 0.0 & 22.5 & 16.0 \\

% \midrule

\multirow{1}{*}{CoT}
& GPT-4o
& 44.0 & 29.0 & 11.7 & 8.3 & 12.5 & 0.0 & 28.0 & 17.0 \\
% & GPT-4o (SC)
% & 45.0 & 34.0 & 11.7 & 5.0 & 15.0 & 0.0 & 29.0 & 18.5 \\

\midrule

\multirow{5}{*}{RAG}
& Claude-3.5-Sonnet
& 27.0 & 13.0 & 23.0 & 6.7 & 7.5 & 0.0 & 22.0 & 8.5 \\
& LLaMA-3.1-70B
& 58.0 & 32.0 & 23.0 & 10.0 & 15.0 & 5.0 & 39.0 & 20.0 \\
& Qwen2.5-70B
& 60.0 & 33.0 & 25.0 & 13.0 & 15.0 & 5.0 & 40.5 & 21.5 \\
& GPT-4o
& 62.0 & 41.0 & 45.0 & 21.7 & 40.0 & 7.5 & 52.0 & 23.0 \\
& o1-preview
& 70.0 & 46.0 & 48.3 & 23.3 & 30.0 & 12.5 & 55.5 & 32.5 \\

\midrule

\multirow{1}{*}{ComfyAgent}
& Qwen2.5-72B
& 63.0 & 38.0 & 25.0 & 18.3 & 20.0 & 5.0 & 43.0 & 25.5 \\

% \midrule

\multirow{1}{*}{SFT+GRPO}
& ComfyUI-R1
& - & - & - & - & - & - & 67.0 & - \\
% \midrule

\multirow{1}{*}{ComfyMind}
& Qwen2.5-72B
& - & 80.0 & - & 63.0 & - & 25.0 & - & 64.0 \\
% \midrule

\multirow{1}{*}{\textbf{Ours}}
& \textbf{ComfySearch-7B}
& \textbf{92.0} & \textbf{80.0}
& \textbf{91.6} & \textbf{76.6}
& \textbf{95.0} & \textbf{42.5}
& \textbf{92.5} & \textbf{71.5} \\

\bottomrule
\end{tabular}
}
\caption{ ComfyBench ~\cite{Xue2024ComfyBenchBL} evaluation of autonomous ComfyUI workflow construction. \%Pass measures whether the generated workflow executes successfully in ComfyUI, and \%Resolve measures whether the executed output satisfies all task requirements. Results are reported by difficulty subset and overall; ``-'' denotes metrics not reported by the corresponding method or not comparable. Higher is better.}
\label{tab:comfybench}
\end{table*}

\paragraph{Tool-feedback RL with GRPO.}
We adopt Group Relative Policy Optimization (GRPO)~\cite{Shao2024DeepSeekMathPT}. For each query, we generate a group of rollouts under the tool-mediated environment(including entropy-adaptive branching under a fixed budget), compute terminal rewards, and update the policy with GRPO (full objective in Appendix~\ref{sec:appendix:grpo}).

\subsection{Hierarchical Terminal Reward}
\label{sec:method:reward}
We use a hierarchical terminal reward that evaluates (i) output format, (ii) trace--workflow consistency, and (iii) structural accuracy against $G^\star$.

\paragraph{$R_f$: format validity.}
We require a well-formed \texttt{<trace>} block (containing \texttt{<node>} and \texttt{<result>} steps) and a parseable \texttt{<workflow>} block. If the format is invalid, we set $R_f=-1$. 

\paragraph{$R_c$: trace--workflow consistency.}
We further require that each node type appearing in the final workflow is mentioned in the trace; otherwise $R_c=-1$.

\paragraph{Structural term.}
Let $\textsc{Types}(G)$ be the set of node types in graph $G$. We use node-type recall:
\begin{equation}
\widehat{R}_s(G_T,G^\star) \triangleq
\frac{|\textsc{Types}(G_T)\cap \textsc{Types}(G^\star)|}{|\textsc{Types}(G^\star)|}-1.
\label{eq:recall}
\end{equation}

\paragraph{Final reward.}
We apply veto-style gates and a graded structural score:
\begin{equation}
R(\tau) =
\begin{cases}
-1, & \text{if } R_f=-1 \text{ or } R_c=-1,\\
\frac{3 + \widehat{R}_s(G_T,G^\star)}{3}, & \text{otherwise.}
\end{cases}
\label{eq:final_reward}
\end{equation}
This provides informative signals for valid trajectories while sharply penalizing invalid outputs.

\section{Experiments}
\subsection{Experiment Settings}
\paragraph{Dataset and Benchmarks.}
We build our training data from Flow-Dataset \cite{huang2025comfygptselfoptimizingmultiagentcomprehensive}, which contains community-shared ComfyUI workflows. Since such workflows are rarely accompanied by aligned natural-language queries, we follow the data construction pipeline in Section~\ref{sec:data:consistency} to obtain query--workflow supervision, including consistency-filtered pairing and validator-grounded reasoning traces. We evaluate ComfySearch on two benchmarks: ComfyBench \cite{Xue2024ComfyBenchBL} measures autonomous workflow construction under strict executability constraints, and GenEval \cite{ghosh2023geneval} evaluates downstream text-to-image compositional accuracy.

\begin{table*}[htbp]
\small
  \centering
  % \caption{Evaluation of T2I generation on GenEval~\cite{ghosh2023geneval}. Obj.: Object. Attr.: Attribution.}
  \resizebox{\linewidth}{!}{
      \begin{tabular}{l|c|cccccc} % 定义 9 列：1 个左对齐列 + 8 个居中列
        \toprule
        \textbf{Method}  & \textbf{Overall} & \textbf{Single Obj.} & \textbf{Two Obj.} & \textbf{Counting} & \textbf{Colors} & \textbf{Position} & \textbf{Attri. Binding} \\
        \midrule
        \multicolumn{8}{l}{\textit{Frozen Text Encoder Mapping Methods}} \\
        SDv1.5~\cite{rombach2022high}                  & 0.43 & 0.97 & 0.38 & 0.35 & 0.76 & 0.04 & 0.06 \\
        SDv2.1~\cite{rombach2022high}                  & 0.50 & 0.98 & 0.51 & 0.44 & 0.85 & 0.07 & 0.17 \\
        SD-XL~\cite{podell2023sdxl}                    & 0.55 & 0.98 & 0.74 & 0.39 & 0.85 & 0.15 & 0.23 \\
        DALLE-2~\cite{ramesh2022hierarchical}          & 0.52 & 0.94 & 0.66 & 0.49 & 0.77 & 0.10 & 0.19 \\
        SD3-Medium~\cite{esser2024scaling}             & 0.74 & 0.99 & 0.94 & 0.72 & 0.89 & 0.33 & 0.60 \\
        \midrule
        \multicolumn{8}{l}{\textit{LLMs/MLLMs Enhanced Methods}} \\
        LlamaGen~\cite{sun2024autoregressive}                   & 0.32 & 0.71 & 0.34 & 0.21 & 0.58 & 0.07 & 0.04 \\
        Chameleon~\cite{team2024chameleon}                      & 0.39 & -    & -    & -    & -    & -    & -    \\
        LWM~\cite{liu2024world}                                 & 0.47 & 0.93 & 0.41 & 0.46 & 0.79 & 0.09 & 0.15 \\
        SEED-X~\cite{ge2024seed}                                & 0.49 & 0.97 & 0.58 & 0.26 & 0.80 & 0.19 & 0.14 \\
        Emu3-Gen~\cite{wang2024emu3}                            & 0.54 & 0.98 & 0.71 & 0.34 & 0.81 & 0.17 & 0.21 \\
        Janus~\cite{wu2024janus}                                & 0.61 & 0.97 & 0.68 & 0.30 & 0.84 & 0.46 & 0.42 \\
        JanusFlow~\cite{ma2024janusflow}                        & 0.63 & 0.97 & 0.59 & 0.45 & 0.83 & 0.53 & 0.42 \\
        Janus-Pro-7B~\cite{chen2025janus}                        & 0.80 & 0.99 & 0.89 & 0.59 & \textbf{0.90} & \textbf{0.79} & 0.66 \\
        GoT~\cite{fang2025got}                                  & 0.64 & 0.99 & 0.69 & 0.67 & 0.85 & 0.34 & 0.27 \\
        \midrule
        \multicolumn{8}{l}{\textit{Workflow Generation Systems}} \\
        ComfyAgent~\cite{Xue2024ComfyBenchBL}  & 0.32  & 0.69 & 0.30 & 0.33 & 0.50 & 0.04 & 0.04 \\
        ComfyMind~\cite{Xue2024ComfyBenchBL}  & 0.80  & 0.95 & 0.88 & \textbf{0.83} & 0.89 & 0.58 & 0.60 \\
        \textbf{ComfySearch}        & \textbf{0.82} & \textbf{1.00} & \textbf{0.91} & 0.79 & 0.86 & 0.65 & \textbf{0.75} \\
        \bottomrule
      \end{tabular}
    }
  \caption{GenEval~\cite{ghosh2023geneval} text-to-image compositional accuracy. For workflow-generation systems, we execute the generated ComfyUI workflows to synthesize images and then evaluate them with GenEval. Obj.: object; Attr.: attribute. Higher is better.}
  \label{tab:t2i-geneval}
\end{table*}

\paragraph{Implementation Details.}
We use Qwen2.5-Coder-7B-Instruct as the backbone of ComfySearch. The SFT stage is trained on 8 NVIDIA A100 (80GB) GPUs for 1 epoch with learning rate $2\times 10^{-5}$ and per-GPU batch size 1. During RL training, we use 4 NVIDIA A100 (80GB) GPUs for 600 steps with learning rate $5\times 10^{-7}$ and total batch size 16. In Eq.~\ref{eq:branch_prob}, we set $\beta=0.2$, $\alpha=0.5$, and $\tau_b=0.5$.

% \noindent\textbf{Implementation Details}
% We use Qwen2.5-Coder-7B-Instruct as the backbone of ComfySearch. The SFT training stage is conducted on 8 NVIDIA A100 GPUs(80G) over 1 epoch, with a learning rate of 2e-5 and a batch size of 1 on each GPU. During RL training, we utilize 4 Nvidia A100 GPUs(80G) to train for a total of 600 steps, with a learning rate of 5e-7 and a total batch size of 16. 
% % In Eq \ref{eq:rl_loss}, the number of group computations $G$ is set to 5. 
% In Eq \ref{eq:branch_prob}, we set $\beta$, $\alpha$ and $\tau$ to 0.2, 0.5, and 0.5, respectively.

\subsection{Autonomous Workflow Construction}

We evaluate autonomous workflow construction on ComfyBench, which contains 200 generation and editing tasks spanning image and video modalities with three difficulty subsets (Vanilla, Complex, and Creative). For each task, the model must synthesize a ComfyUI workflow that (i) executes successfully and (ii) satisfies the user instruction. ComfyBench reports two metrics: executability (\%Pass), indicating whether the workflow runs without execution errors, and task success (\%Resolve), indicating whether the executed output meets all task requirements.

As shown in Table~\ref{tab:comfybench}, ComfySearch achieves strong overall performance with 92.5 \%Pass and 71.5 \%Resolve, substantially outperforming zero-/few-shot prompting and retrieval-augmented baselines. Compared to the learning-based workflow generator ComfyUI-R1, ComfySearch improves the overall \%Pass from 67.0 to 92.5, indicating markedly more reliable executability. Relative to ComfyAgent, ComfySearch improves both \%Pass and \%Resolve across all difficulty levels. Compared to the state-of-the-art ComfyMind, ComfySearch matches its \%Resolve on Vanilla (80.0) while achieving large gains on Complex (76.6 vs.\ 63.0) and Creative (42.5 vs.\ 25.0), resulting in a higher overall \%Resolve (71.5 vs.\ 64.0). We attribute these gains to reasoning-as-action workflow construction: state-aware validation with in-place repair prevents early structural errors from compounding, and entropy-adaptive branching enables targeted exploration at high-uncertainty decision points.

\subsection{Text-to-Image Generation}

We report downstream text-to-image performance on GenEval. GenEval evaluates compositional alignment across six dimensions, including single/multiple objects, counting, color accuracy, spatial positioning, and attribute binding. For workflow-generation systems, we execute the generated ComfyUI workflows to synthesize images and then compute GenEval scores.

As shown in Table~\ref{tab:t2i-geneval}, ComfySearch achieves an overall GenEval score of 0.82, outperforming ComfyAgent (0.32) by a large margin and improving over SD3-Medium (0.74) among frozen text-encoder mapping methods. ComfySearch also slightly improves over ComfyMind (0.80) and is competitive with strong multimodal generation systems such as Janus-Pro-7B (0.80), achieving particularly strong performance on composition-sensitive dimensions (e.g., attribute binding).

\begin{figure*}[htbp]
  \centering
  \includegraphics[width=1.0\textwidth]{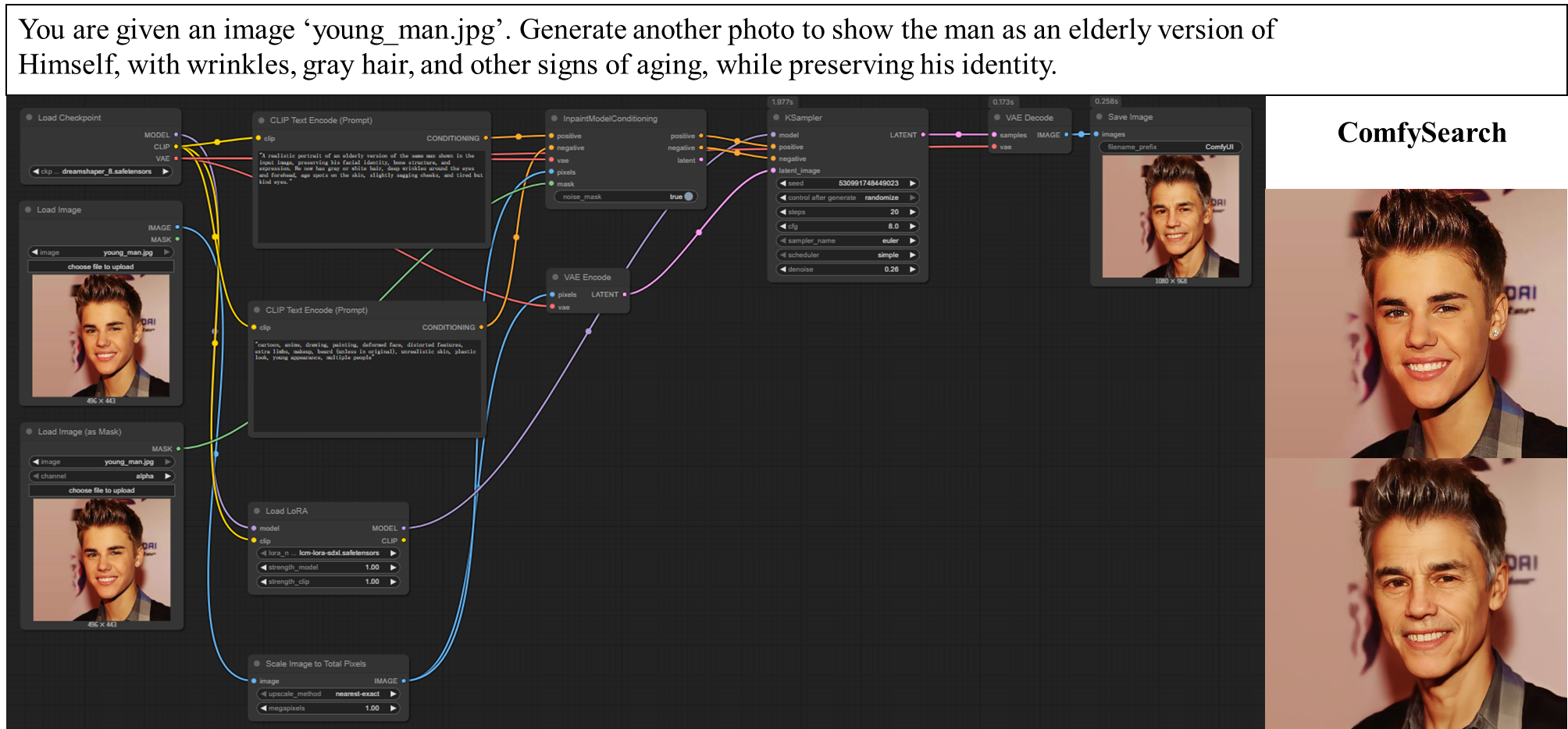}
  \vspace{-0.5em} % 调整图与标题之间的间距
  \caption{Examples generated by ComfySearch.}
  \label{fig:case}
\end{figure*}

\subsection{Qualitative Results}
\begin{figure*}[t]
  \centering
  \includegraphics[width=1.0\textwidth]{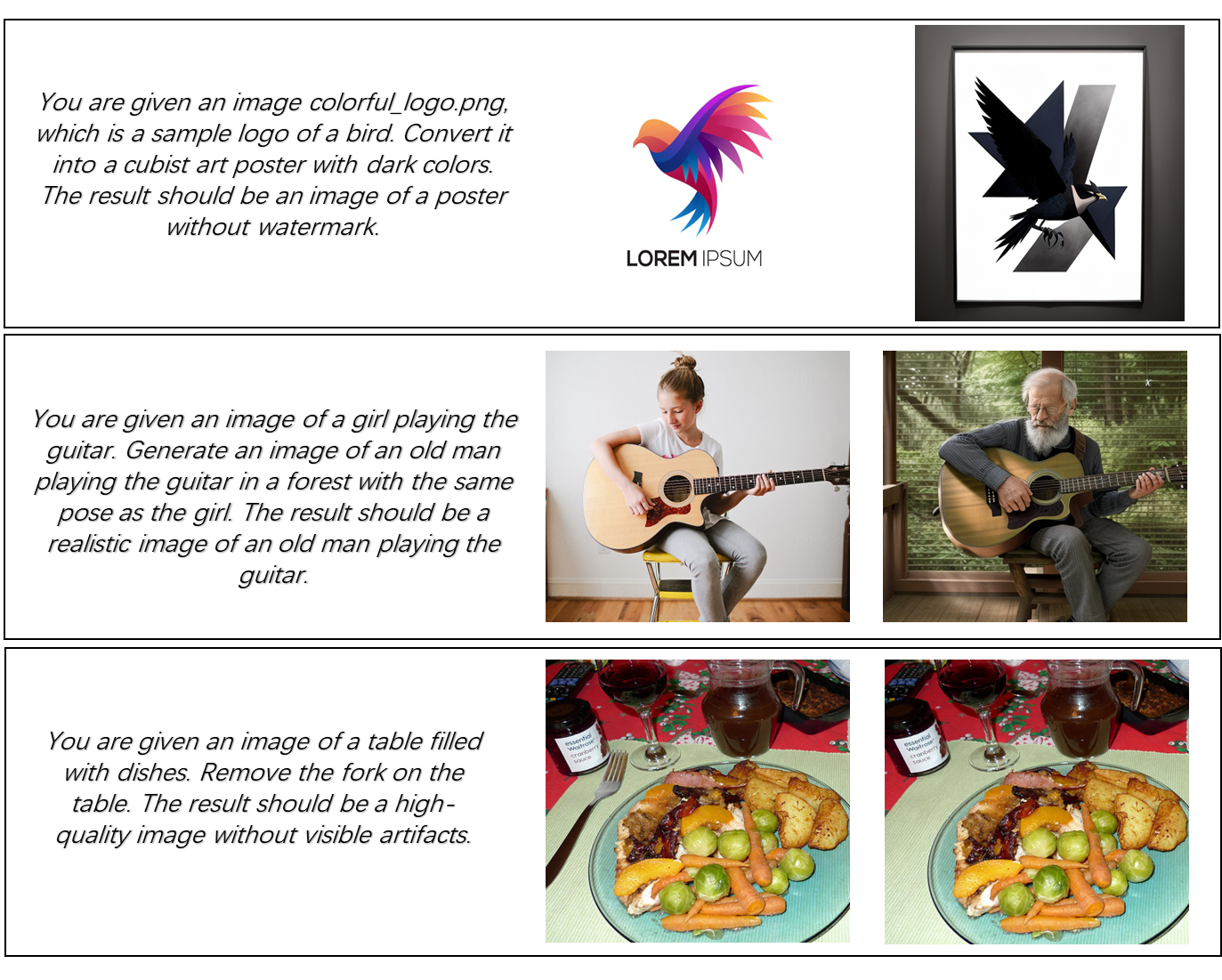}
  \caption{Qualitative results showing the performance of our method on various tasks.}
  \label{fig:result}
\end{figure*}

Figure~\ref{fig:case} shows a representative example workflow generated by ComfySearch from a user query, illustrating that ComfySearch can interpret the instruction, construct a valid executable workflow, and complete the task successfully. Additional qualitative comparisons, including execution outputs and side-by-side results against ComfyMind, are provided in Figure \ref{fig:result} and Appendix~\ref{apd:qualitative}.

\begin{table}[htbp]
  \centering
  \resizebox{\linewidth}{!}{
    \begin{tabular}{l|cccc}
      \toprule
      \textbf{Methods} 
      & \begin{tabular}[c]{@{}c@{}}\textbf{Prompt}\\ Tokens\end{tabular}
      & \begin{tabular}[c]{@{}c@{}}\textbf{Completion}\\ Tokens\end{tabular}
      & \begin{tabular}[c]{@{}c@{}}\textbf{Wall-clock}\\ Time (s)\end{tabular}
      & \textbf{Resolve (\%)} \\
      \midrule
      ComfySearch & 9,332 & 24,895 & 145 & 100 \\
      ComfyMind   & 145,532 & 11,400 & 185 & 100 \\
      \bottomrule
    \end{tabular}
  }
  \caption{Inference efficiency comparison between ComfySearch and ComfyMind on 20 ComfyBench instances filtered to those solved by both methods. We report total prompt tokens, completion tokens, and end-to-end wall-clock time aggregated over the 20 instances; Resolve denotes the task success rate on this filtered subset.}
  \label{tab:efficiency}
\end{table}

\subsection{Efficiency Analysis}

We randomly sample 20 ComfyBench instances and compare inference efficiency between ComfySearch and ComfyMind. To focus on efficiency rather than failure cases, we restrict the analysis to instances that are successfully solved by both methods; consequently, both methods achieve 100\% \%Resolve on this filtered subset (Table~\ref{tab:efficiency}). Under this controlled setting, ComfySearch uses substantially fewer total tokens and achieves lower end-to-end wall-clock time than ComfyMind. This advantage is consistent with ComfyMind's iterative tree-based composition and replanning, which can incur large prompt overhead, whereas our ComfySearch performs reasoning-as-action editing with immediate validation/repair signals and selective exploration.

\begin{table}[htbp]
  \centering
  \resizebox{\linewidth}{!}{
    \begin{tabular}{l|cccccccc}
      \toprule
      \multirow{2}{*}{\textbf{Methods}} & \multicolumn{2}{c}{\textbf{Vanilla}} & \multicolumn{2}{c}{\textbf{Complex}} & \multicolumn{2}{c}{\textbf{Creative}}\\
      \cmidrule(lr){2-3} \cmidrule(lr){4-5} \cmidrule(lr){6-7}
      & \%Pass & \%Resolve & \%Pass & \%Resolve & \%Pass & \%Resolve \\
      \midrule
      \textbf{ComfySearch}       & \textbf{92.0}  & \textbf{80.0}  & \textbf{91.6}  &\textbf{76.6} & \textbf{95.0}  & \textbf{42.5}\\   
      SFT   & 79.0  & 60.0  & 60.0  & 30.0  & 72.5  & 25.0\\      
      SFT+GRPO    & 80.0  & 65.0  & 66.6  & 38.0 & 87.5  & 30.0 \\    
      SFT+GRPO+VAL & 88.0  & 74.0  & 88.3  & 73.0 & 92.5  & 37.5 \\    
      RAG+SFT+GRPO & 86.0  & 68.0  & 70.0  & 55.0 & 87.5  & 35.0 \\   
      \bottomrule
    \end{tabular}
  }
    \caption{Ablation results of our ComfySearch components on ComfyBench. \emph{VAL} corresponds to the state-aware validation and in-place repair module (C1). Higher is better.}
    \label{tab:ablation}
\end{table}

\subsection{Ablation Study}
\paragraph{Method Ablation.}
Table~\ref{tab:ablation} ablates key training and inference components in ComfySearch. Starting from SFT, adding RL (SFT+GRPO) consistently improves both \%Pass and \%Resolve. Incorporating \emph{state-aware validation} (denoted as \emph{VAL}) on top of SFT+GRPO yields further gains, especially on Complex and Creative tasks, demonstrating the importance of online structural correctness (C1). The full ComfySearch system achieves the best overall performance, suggesting that uncertainty-guided entropy-adaptive branching complements validation-guided editing in long-horizon workflow construction (C2). We also compare to a retrieval-augmented variant (RAG+SFT+GRPO), which underperforms ComfySearch on the harder subsets, indicating that execution-grounded exploration is more effective than relying on static retrieval alone.

\begin{table}[htbp]
  \centering
  \resizebox{\linewidth}{!}{
    \begin{tabular}{l|cccccccc}
      \toprule
      \multirow{2}{*}{\textbf{Datasets}} & \multicolumn{2}{c}{\textbf{Vanilla}} & \multicolumn{2}{c}{\textbf{Complex}} & \multicolumn{2}{c}{\textbf{Creative}}\\
      \cmidrule(lr){2-3} \cmidrule(lr){4-5} \cmidrule(lr){6-7}
      & \%Pass & \%Resolve & \%Pass & \%Resolve & \%Pass & \%Resolve \\
      \midrule
      \textbf{Ours}       & \textbf{92.0}  & \textbf{80.0}  & \textbf{91.6}  &\textbf{76.6} & \textbf{95.0}  & \textbf{42.5}\\   
      w/o Consistency   & 85.0  & 66.0  & 80.0  & 63.0  & 92.5  & 30.0\\      
      w/o Reasoning    & 85.0  & 68.0  & 78.3  & 60.0 & 87.5  & 32.5 \\    
      \bottomrule
    \end{tabular}
  }
    \caption{Ablation of training-data construction (\S\ref{sec:data:consistency}) on ComfyBench. ``w/o Consistency'' removes consistency-filtered pairing, and ``w/o Reasoning'' removes validator-grounded reasoning traces. Higher is better.}
    \label{tab:dataset}
\end{table}

\paragraph{Data Construction Ablation.}
To evaluate the data construction components in \S\ref{sec:data:consistency}, we remove either consistency-filtered pairing (\emph{w/o Consistency}) or validator-grounded reasoning traces (\emph{w/o Reasoning}). As shown in Table~\ref{tab:dataset}, removing either component consistently degrades performance across all difficulty levels. In particular, removing reasoning traces leads to notable drops on Complex and Creative tasks, highlighting the importance of intermediate supervision for robust long-horizon workflow construction.

\section{Conclusion}
We present ComfySearch, an reasoning-as-action framework for autonomous ComfyUI workflow generation. We formulate the process as a Markov Decision Process (MDP) with two core innovations: (1) State-aware validation and in-place repair to ensure online structural correctness, preventing error propagation; and (2) Entropy-adaptive branching to navigate long-horizon uncertainty. Optimized via GRPO with hierarchical rewards, ComfySearch enables robust exploration beyond static planning. Experiments on ComfyBench and GenEval demonstrate that our 7B model significantly outperforms baselines in executability, task resolution, and generalization for complex creative workflows.

\section*{Limitations}
ComfySearch relies on explicit validators and curated execution-verified supervision, which may limit performance when encountering entirely unseen components or rapidly evolving node libraries. Although entropy-adaptive branching mitigates exhaustive search, the exploration cost can still grow for highly ambiguous or very large workflow spaces. Moreover, our evaluation focuses on ComfyUI-centric benchmarks, and validating the generality of this execution-grounded framework across other workflow systems and domains remains an important direction for future work.

\bibliography{custom}

\clearpage

\appendix
\section{An example of SFT data}
\label{apd:sft_data}
After our data construction pipeline, we obtained 4,268 curated query–workflow pairs for model training.
\begin{tcolorbox}[notitle, sharp corners, breakable,  colback=white, 
       boxrule=3pt, boxsep=0.5pt, 
       title={sample}]\label{box:operator-profile}
       \footnotesize
       {\fontfamily{pcr}\selectfont
\begin{lstlisting}
"instruction": "You are a reasoning assistant specialized in building valid ComfyUI workflows step by step. For a given creative or multimodal generation task, you must first reason which nodes are required and how they connect to form a valid workflow. The top-level structure of your output must be exactly:
 1. <thinking></thinking>  enclosing the entire reasoning process. Inside <thinking>, include all reasoning, and for each proposed node, wrap its single line of valid Python code inside <node></node> for validation. Each <node> can only contain one line of code. After a <node>, if a <result>...</result> section appears, it shows the validation feedback. If no <result> appears, the node has passed validation.
 2. <workflow></workflow>  enclosing the final workflow. This should contain only the validated Python code lines in the proper order, one line per node. Do NOT include any <thinking>, <node>, <result>, or any other tags inside <workflow>.
Always follow this structure: the top-level output must start with <thinking>...</thinking> followed immediately by <workflow>...</workflow>, with no other wrapping or nested top-level elements. Each Python line in <workflow> must be valid and executable.",

"query": "You are given an image `low_res_image.png`, which is a low-resolution photo of a cityscape at sunset. Enhance the resolution of the image using the best available upscaling method to create a high-quality, detailed version of the same cityscape. The result should be a high-resolution image that maintains the original colors and details.",

"Reasoning": 
<thinking>The user's goal is to upscale a low-resolution cityscape image to high-resolution while preserving colors and details. The key requirements are high-quality upscaling, color fidelity, and detail preservation. The workflow uses a Stable Diffusion-based approach with text conditioning for guidance.<node><result>...The CheckpointLoaderSimple node is required to load the foundational model components (diffusion model, CLIP encoder, and VAE) necessary for the upscaling process, ensuring compatibility with Stable Diffusion-based methods.Data flows from loading the model and image, through text conditioning, to the upscaling process, and finally to saving the result. Each node is critical for achieving the desired outcome.
</thinking>


<workflow>
checkpointloadersimple_0_model, checkpointloadersimple_0_clip, checkpointloadersimple_0_vae = CheckpointLoaderSimple()...
    SaveImage(images=
    ultimatesdupscale_0_image)
</workflow>"
\end{lstlisting}
}
\end{tcolorbox}

\section{GRPO}
\label{sec:appendix:grpo}

For each query $Q$, we sample a group of $K$ trajectories 
$\{\tau^{(k)}\}_{k=1}^{K}$ (generated by the validate--repair transition and optional branching), 
and obtain their terminal rewards $\{R(\tau^{(k)})\}$.
We compute a group-relative advantage:
\begin{equation}
\hat{A}^{(k)} \triangleq 
R(\tau^{(k)}) - \frac{1}{K} \sum_{j=1}^{K} R(\tau^{(j)}).
\label{eq:grpo_adv}
\end{equation}

The GRPO objective with KL regularization is:
\begin{align}
\max_{\theta} \; 
& \mathbb{E}_{k,t} \Bigl[ \hat{A}^{(k)} \log 
  \pi_{\theta}\bigl(a_t^{(k)} \mid s_t^{(k)}\bigr) \Bigr] \notag \\
& - \lambda_{\mathrm{KL}} \, 
\mathrm{KL}\bigl(\pi_{\theta} \,\|\, \pi_{\mathrm{ref}}\bigr),
\label{eq:grpo_obj}
\end{align}
where $\pi_{\mathrm{ref}}$ is the reference policy (initialized from SFT) 
and $\lambda_{\mathrm{KL}}>0$ controls update stability.

\section{Notation.}
\label{notations}
We conclude the commonly used notations for reference.
\begin{itemize}
\item \textbf{$w_g^\star$}: The canonical workflow selected for a query group $Q_g$ by maximizing the sum of correctness scores $s(w, Q)$ over workflows $W_g$ that succeed for at least one query in the group.
\item \textbf{$s_t$}: The state in the MDP formulation, consisting of the query $Q$, current partial workflow $G_t$, and history $h_t$ of recent diagnostics and constraints.
\item \textbf{$\Delta_t$}: An atomic graph edit parsed deterministically from the policy's textual action $a_t$ via the parser $\Phi$.
\item \textbf{$G_{t+1}, y_t, m_t$}: The tool-mediated transition output from VALIDATE, where $y_t \in \{0, 1\}$ indicates acceptance and $m_t$ is the diagnostic message (empty if accepted).
\item \textbf{$h_t$}: Summarizes recent diagnostic information and accumulated constraints. Its presence ensures that the construction process in tool-mediated environments possesses the Markov property.
\item \textbf{$y_t$}: The state-aware validation acceptance indicator, composing intrinsic validity INT and composability COMP.
\item \textbf{$m_t$}: The diagnostic message specifying the violated constraint if validation fails.
\item \textbf{$a_t^{\mathrm{rep}}$}: The repaired action proposed by the policy conditioned on the diagnostic $m_t$ if validation fails.
\item \textbf{$\Delta_t^{\mathrm{rep}}$}: The repaired edit parsed from the repaired action $a_t^{\mathrm{rep}}$.
\item \textbf{$\tilde{A}_t$}: The constrained candidate action set at step $t$.
\item \textbf{$p_t(a)$}: The policy probability for action $a$ given state $s_t$.
\item \textbf{$H_t$}: The normalized entropy of the policy over candidate actions $\tilde{A}_t$, measuring step-level uncertainty.
\item \textbf{$\Delta H_t$}: The entropy increase from the previous step, used to trigger branching.
\item \textbf{$P_t^{\mathrm{branch}}$}: The branching probability based on entropy increase, using sigmoid with parameters $\alpha$ and $\beta$.
\item \textbf{$\tau_b$}: The branching threshold; if $P_t^{\mathrm{branch}} > \tau_b$, spawn an additional branch.
\item \textbf{$\mathcal{L}_{\mathrm{SFT}}(\theta)$}: The supervised fine-tuning loss over training tuples of queries $Q$, teacher traces $T$, and target workflows $G^\star$.
\item \textbf{$R_f$}: The format validity reward component, requiring well-formed trace and workflow blocks.
\item \textbf{$R_c$}: The trace-workflow consistency reward component, ensuring node types in the workflow appear in the trace.
\item \textbf{$R_s(G_T, G^\star)$}: The node-type recall for structural accuracy, measuring overlap in node types between generated $G_T$ and target $G^\star$.
\item \textbf{$R(\tau)$}: The hierarchical terminal reward, combining format $R_f$, consistency $R_c$, and structural $R_s$ with veto penalties for invalid outputs.
\end{itemize}

\section{Qualitative Results}
\label{apd:qualitative}
\begin{figure*}[!htbp]
  \centering
  \includegraphics[width=1.0\textwidth]{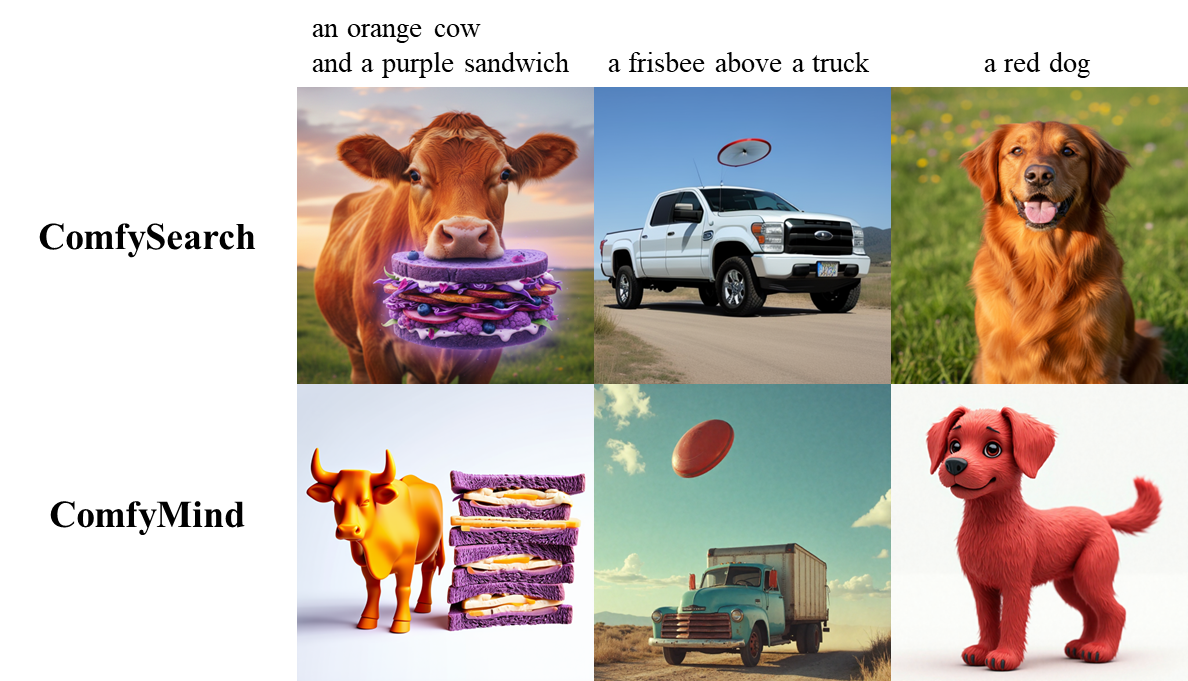}
  \caption{Qualitative comparison between our method and ComfyMind}
  \label{cop}
\end{figure*}

Figure~\ref{cop} shows a qualitative comparison between our method and ComfyMind, illustrating that our method achieves better task completion and higher-quality outputs.

\section{Evaluation}
\label{apd:evaluation}
We follow ComfyBench's evaluation—using GPT-4o for assessment—with text-to-image as an example.
\begin{tcolorbox}[notitle, sharp corners, breakable,  colback=white, 
       boxrule=3pt, boxsep=0.5pt, 
       title={Evaluation prompt}]\label{box:operator-profile}
       \footnotesize
       {\fontfamily{pcr}\selectfont
\begin{lstlisting}
'''
You are an expert in image and video generation, familiar with the latest tasks and techniques. You are capable of understanding the task instruction, analyzing the generation result, and providing an accurate evaluation. Now you are evaluating the result of a text-to-image generation task. You should be tolerant to the quality of the generation result, and focus on the consistency with the instruction.

The task instruction is described as: {instruction}

The given image is the generation result, with an actual resolution of {result_resolution}.

First, analyze whether the generation result meets each key point in the instruction. Enclose your analysis in the <analysis> tag. For example: <analysis>There is a cat in an astronaut suit, which is consistent with the instruction. The wall is white, which is different from the "green wall" in the instruction.</analysis>.

Then, provide a final judgment of whether the generation result complies with the instruction. The judgment should either be "True" or "False". Enclose your judgment in the <judgment> tag. For example: <judgment>False</judgment>.
'''
\end{lstlisting}
}
\end{tcolorbox}

\end{document}